\documentclass[letterpaper,10pt,conference]{ieeeconf}  

\IEEEoverridecommandlockouts

\usepackage{xparse}

\usepackage{amsmath} 
\usepackage{amssymb}
\usepackage{amsthm}
\usepackage{amsfonts} 
\usepackage{color}

\usepackage{graphicx}
\graphicspath{{figs/}}
\usepackage{calc}

\IEEEtriggeratref{20}

\usepackage[ruled,vlined]{algorithm2e}
\usepackage{etoolbox}%
\providetoggle{beginFlag}
\settoggle{beginFlag}{true}

\usepackage{hyperref}

\usepackage{cleveref}
\crefformat{equation}{(#2#1#3)}
\Crefname{figure}{Fig.}{Figs.}
\crefname{figure}{Fig.}{Figs.}
\crefformat{figure}{#2Fig.~#1#3}
\Crefformat{figure}{#2Fig.~#1#3}
\crefformat{table}{#2Table~#1#3}
\Crefformat{table}{#2Table~#1#3}
\crefformat{section}{#2Sec.~#1#3}
\Crefformat{section}{#2Sec.~#1#3}
\crefformat{problem}{#2Problem~#1#3}
\Crefformat{problem}{#2Problem~#1#3}
\crefname{appsec}{Appendix}{Appendices}

\newtheorem{assumption}{Assumption}

\usepackage{url}
\usepackage{graphicx}

\usepackage{array}
\usepackage{longtable}
\usepackage{multirow}
\usepackage{float}
\usepackage{caption}
\usepackage{mathrsfs}
\usepackage{listings}
\usepackage{rotating}
\usepackage{overpic}

\usepackage{booktabs}

\usepackage{xparse}

\newcommand{\lrss}[5]{%
	\setbox1=\hbox{\ensuremath{^{#1}}}%
	\setbox2=\hbox{\ensuremath{_{#2}}}%
	\setbox5=\hbox{\ensuremath{#5}}%
	\setbox6=\hbox{\ensuremath{^{#1#3}}}%
	\setbox7=\hbox{\ensuremath{_{#2#4}}}%
	\setbox8=\hbox{\ensuremath{^{#3}}}%
	\setbox9=\hbox{\ensuremath{_{#4}}}%
	\hspace{\ifnum\wd1>\wd2\wd1\else\wd2\fi}%
	\ensuremath{\copy5%
		^{\hspace{-\wd1}\hspace{\wd1}\hspace{\wd8}%
			\hspace{-\wd6}\hspace{-\wd5}#1\hspace{\wd5}#3}%
		_{\hspace{-\wd2}\hspace{\wd2}\hspace{\wd9}%
			\hspace{-\wd7}\hspace{-\wd5}#2\hspace{\wd5}#4}%
}}

\NewDocumentCommand\bbm{}{ \begin{bmatrix} }
\NewDocumentCommand\ebm{}{ \end{bmatrix} }
\NewDocumentCommand\Vector{m}{ \boldsymbol{\mathbf{#1}} }
\NewDocumentCommand\Matrix{m}{ \boldsymbol{\mathbf{#1}} }

\NewDocumentCommand\Norm{m}{\left\Vert#1\right\Vert }

\NewDocumentCommand\Transpose{}{\mathsf{T}}
\NewDocumentCommand\Vech{}{\mathrm{vech}}

\NewDocumentCommand\Real{}{ \mathbb{R} }

\NewDocumentCommand\Natural{}{ \mathbb{N} }

\NewDocumentCommand\CoordinateFrame{m}{ \underrightarrow{\Matrix{\mathcal{F}}}_{#1} }

\NewDocumentCommand\PositionSymbol{}{ t }
\NewDocumentCommand\RotationSymbol{}{ R }

\NewDocumentCommand\Rot{mm}{ {\lrss{#2}{}{#1}{}{\Matrix{\RotationSymbol}}} }

\NewDocumentCommand\Pos{mmm}{ \lrss{#2}{}{#1}{#3}{\Vector{\PositionSymbol}} }

\NewDocumentCommand\Part{m}{ r_{#1} }

\NewDocumentCommand\Skew{m}{{\left[#1\right]\!\!_{_\times}}}

\NewDocumentCommand\Mass{}{ m }

\NewDocumentCommand\InertiaMatrix{}{ \Matrix{J} }

\NewDocumentCommand\Gravity{}{ g }
\NewDocumentCommand\Force{}{ f }
\NewDocumentCommand\Torque{}{ \tau }
\NewDocumentCommand\Position{}{ p }

\NewDocumentCommand\ObjectFrame{}{ \CoordinateFrame{b} }
\NewDocumentCommand\WorldFrame{}{ \CoordinateFrame{w} }
\NewDocumentCommand\SensorFrame{}{ \CoordinateFrame{s} }

\NewDocumentCommand\DataMatrix{}{ \Matrix{A} }
\NewDocumentCommand\RedModel{}{ \Matrix{\widetilde{A}} }

\NewDocumentCommand\TextCOM{}{COM }
\NewDocumentCommand\TextCOMMA{}{COM}

\title{\LARGE \bf The Sum of Its Parts: Visual Part Segmentation for\\ Inertial Parameter Identification of Manipulated Objects}

\author{Philippe Nadeau, Matthew Giamou, and Jonathan Kelly$^\ddagger$
	\thanks{All authors are with the STARS Laboratory at the University of Toronto Institute for Aerospace Studies, Toronto, Ontario, Canada. {\tt\footnotesize <firstname>.<lastname>@robotics.utias.utoronto.ca}}
	\thanks{$^\ddagger$Jonathan Kelly is a Vector Institute Faculty Affiliate. This research was supported in part by the Canada Research Chairs program.}}

\begin{document} 
	
	\maketitle 
	\thispagestyle{empty}
	\pagestyle{empty}
	
	\begin{abstract}
	To operate safely and efficiently alongside human workers, collaborative robots (cobots) require the ability to quickly understand the dynamics of manipulated objects.
	However, traditional methods for estimating the full set of inertial parameters rely on motions that are necessarily fast and unsafe (to achieve a sufficient signal-to-noise ratio).
	In this work, we take an alternative approach: by combining visual and force-torque measurements, we develop an inertial parameter identification algorithm that requires slow or ``stop-and-go'' motions only, and hence is ideally tailored for use  around humans.
	Our technique, called Homogeneous Part Segmentation (HPS), leverages the observation that man-made objects are often composed of distinct, homogeneous parts.
	We combine a surface-based point clustering method with a volumetric shape segmentation algorithm to quickly produce a part-level segmentation of a manipulated object; the segmented representation is then used by HPS to accurately estimate the object's inertial parameters.
	To benchmark our algorithm, we create and utilize a novel dataset consisting of realistic meshes, segmented point clouds, and inertial parameters for 20 common workshop tools.
	Finally, we demonstrate the real-world performance and accuracy of HPS by performing an intricate `hammer balancing act' autonomously and online with a low-cost collaborative robotic arm.
	Our code and dataset are open source and freely available.
	\end{abstract}

	\section{Introduction}
	
	In order to operate effectively and safely, robots require an understanding of the dynamics of their interactions with the environment.
	For collaborative robots (cobots) that are designed to work alongside people, the ability to infer the inertial parameters of manipulated objects is particularly important.
	An object's inertial parameters include its mass, centre of mass (\TextCOMMA), and moments of inertia, all of which are used to perform dynamic manipulation~\cite{mason_dynamic_1993}.
	Force-torque sensors (FT) located in the joints or at the wrist of a robot arm can be used to determine the inertial parameters of a manipulated object.
	However, sensor noise limits parameter identification accuracy, especially for cobots that must (should) move within the safe velocity limits defined by ISO standards 10218 \cite{noauthor_robots_nodate} and 15066 \cite{noauthor_robots_nodate-1}.
	These slow motions lead to lower signal-to-noise ratios in force-torque data, prohibiting the use of inertial parameter estimation techniques appropriate for large, fast-moving industrial robots~\cite{nadeau_fast_2022}. 
	In this paper, we propose an alternative identification method that relaxes the need for accurate kinematics estimates but can nonetheless determine the complete set of inertial parameters, under some mild assumptions.
	
	\begin{figure}[t]
		\centering
		\vspace*{2mm}
		\setlength{\fboxsep}{0pt}%
		\setlength{\fboxrule}{0.75pt}%
		\fbox{\includegraphics[width=1\linewidth-1.5pt]{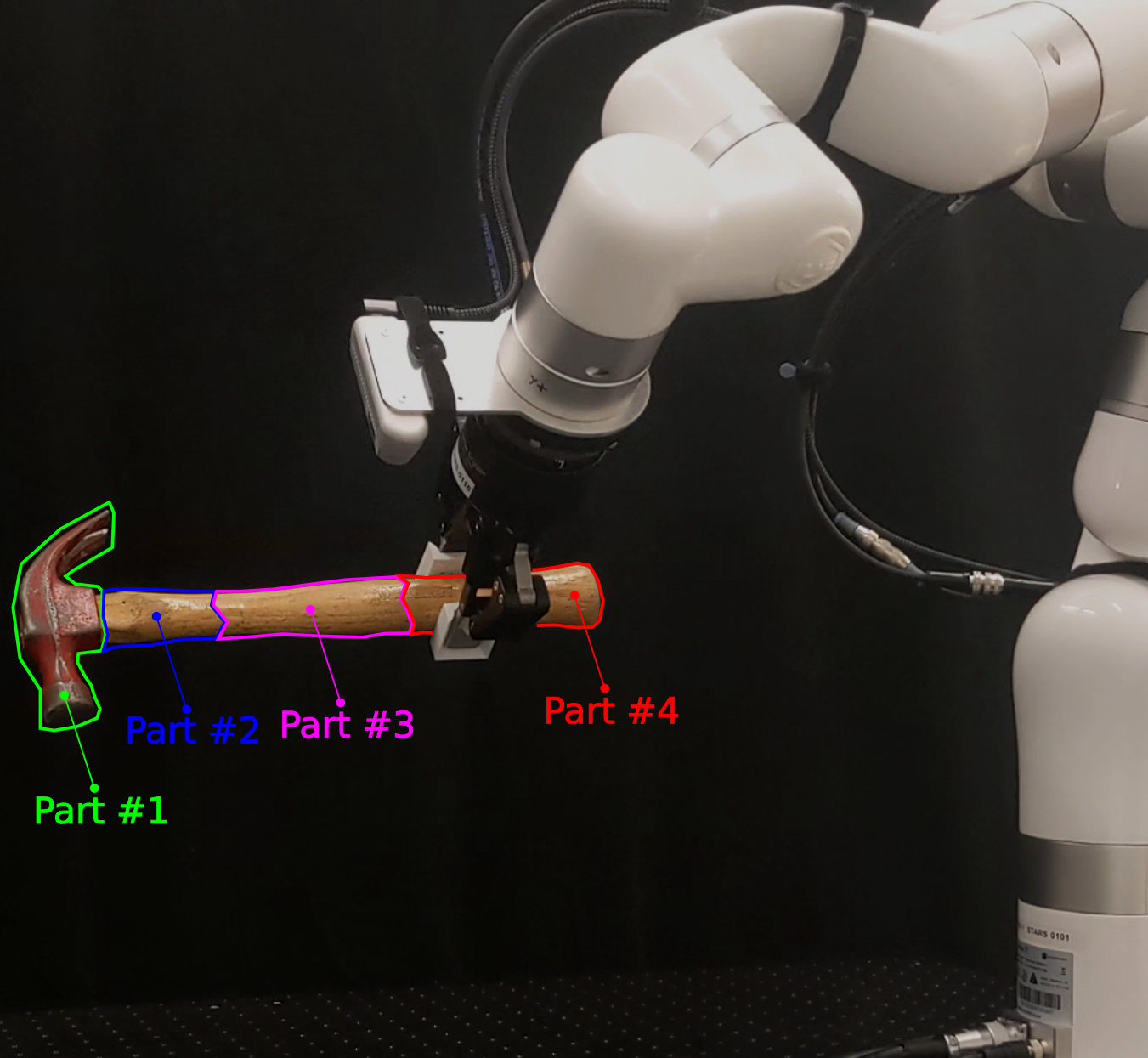}}
		\caption{The shape of a manipulated object is segmented into parts that are assumed to be homogeneous. By identifying the mass of each part, the full set of inertial parameters can be estimated from safer ``stop-and-go'' motions.}
		\label{fig:gladyswieldingobject}
		\vspace{-5mm}
	\end{figure}
	
	Humans are able to easily infer inertial properties through \emph{dynamic touch}~\cite{pagano_eigenvectors_1992} and cobots would benefit from a similar capability~\cite{lukos_choice_2007, hamrick_inferring_2016}.\textsc{}
	Additionally, humans combine proprioception with vision, suggesting that complementary perceptual modalities can speed up the identification process.
	Our technique uses information obtained from an {RGB-D} camera to improve the speed and accuracy of inertial parameter identification.
	We make the key observation that man-made objects are usually built from a few parts having approximately homogeneous densities. 	
	\begin{assumption}[Homogeneous Density of Parts] \label{assum:Homogeneous}
		An object is made of $N$ parts $\{r_1,\dots, r_N\}$, each with a shape $V_j \subset \Real^{3}$. The mass density is such that
		$\forall\,\Vector{\Position}\in V_j,\,\rho(\Vector{\Position}) = \rho_j$
		where $\rho_j>0$ is part $r_j$'s constant density.
	\end{assumption}
	\noindent Critically, \Cref{assum:Homogeneous} can be used to simplify the identification problem by exploiting measurements of the shape of a manipulated object.
	This is accomplished by noting that, once a cobot has determined the shape of an object, the object's inertial parameters depend solely on the mass of each of its homogeneous parts. 
	We refer to our overall approach as \emph{Homogeneous Part Segmentation} (HPS).\footnote{We provide code, our complete simulation dataset, and a video showcasing our algorithm at \url{https://papers.starslab.ca/part-segmentation-for-inertial-identification/}.} 
	Our main contributions are:
	\begin{itemize}
		\item a formulation of inertial parameter identification that incorporates \Cref{assum:Homogeneous};
		\item a method combining the algorithms proposed in \cite{attene_hierarchical_2008} and \cite{lin_toward_2018} to improve part segmentation speed;
		\item a dataset of models of 20 common workshop tools with 3D meshes, point clouds, ground truth inertial parameters, and ground truth part-level segmentation for each object; and
		\item experiments highlighting the benefits of HPS when compared to two benchmark algorithms.
	\end{itemize}
	We undertake a series of simulation studies with our novel dataset and carry out a real-world experiment to assess the performance of the proposed algorithm on a real cobot. We show that, under noisy conditions, our approach produces more accurate inertial parameter estimates than competing algorithms that do not utilize shape information. 
	
	\vspace{2mm}
	\section{Related Work}
	\label{sec:RelatedWork}
	
	This section provides a brief overview of work related to the two main algorithmic components of our system: inertial parameter or load identification and part-level object segmentation.
	A more thorough survey of inertial parameter identification for manipulation is provided by Golluccio et al.\ in \cite{golluccio_robot_2020}.
	
	\subsection{Load Identification}
	
	In \cite{atkeson_estimation_1986}, a least squares method is applied to determine the inertial parameters of a load manipulated by a robot arm.
	The authors of \cite{atkeson_estimation_1986} underline that poor performance can be caused by the low signal-to-noise ratios in the data used for the regression.
	A recursive total least squares (RTLS) approach is proposed in \cite{kubus_-line_2008} to account for noise in the regressor matrix.
	However, an experimental evaluation in \cite{farsoni_real-time_2018} of RTLS on multiple datasets reports large estimation errors, demonstrating that RTLS has limited utility with noisy sensors.
	In this work, we also minimize a least squares cost, but we employ a unique formulation that is well-suited to data gathered by a slower-moving cobot, for example. 
	
	Without appropriate constraints on regression variables, many identification algorithms find unphysical solutions~\cite{sousa_physical_2014, traversaro_identification_2016}.
	Sufficient conditions for the physical consistency of the identified parameters are stated in \cite{traversaro_identification_2016}, where a constrained optimization problem is formulated that guarantees the validity of the solution.
	Similarly, physical consistency is enforced through linear matrix inequalities as part of the method proposed in \cite{wensing_linear_2017}. Geodesic distance approximations from a prior solution are used in \cite{lee_geometric_2019} to regularize the optimization problem without introducing very long runtimes.
	In this work, we 
	\emph{implicitly} enforce the physical consistency of estimated inertial parameters by discretizing the manipulated object with point masses \cite{ayusawa_identification_2010, nadeau_fast_2022}. 
	Fixed-sized voxels can also be used to achieve the same effect~\cite{song_probabilistic_2020}.
	
	Finally, the authors of \cite{sundaralingam_-hand_2021} augment the traditional force-torque sensing system with tactile sensors to estimate the inertial parameters and the friction coefficient of a manipulated object. %
	Our contribution similarly makes use of an additional sensing modality in the form of a camera. 
	
	\subsection{Part-Level Object Segmentation}
	
	Although there is no formal definition of a ``part" of a manipulated item, humans tend to follow the so-called \emph{minima rule}: the decomposition of objects into approximately convex contiguous parts bounded by negative curvature minima \cite{hoffman_parts_1984}.
	This rule has provided inspiration for many part segmentation methods reviewed in \cite{rodrigues_part-based_2018}, which benchmarks several techniques on the Princeton dataset \cite{chen_benchmark_2009} and divides approaches into surface-based, volume-based, and skeleton-based algorithms.
	Surface-based and volume-based mesh segmentation algorithms are reviewed in \cite{shamir_survey_2008}, highlighting a tradeoff between segmentation quality and the number of parts produced by the algorithm.
	Skeleton-based segmentation methods for 3D shapes, which capture structural information as a lower-dimensional graph (i.e., the shape's \emph{skeleton}), are reviewed in \cite{tagliasacchi_3d_2016}.
	The approach in \cite{lin_seg-mat_2020}, which is based on the medial axis transform~\cite{li_q-mat_2015}, exploits prior knowledge of an object's skeleton to produce a \emph{top-down} segmentation about 10 times faster than \emph{bottom-up} methods that use local information only. 
	
	The method of Kaick et al. \cite{kaick_shape_2014} also segments incomplete point clouds into approximately convex shapes, but is prohibitively slow for manipulation tasks, with an average computation time of 127 seconds on the Princeton benchmark \cite{chen_benchmark_2009} as reported in \cite{lin_seg-mat_2020}.
	In contrast, learning-based methods for part segmentation can struggle to generalize to out-of-distribution shapes but are usually faster than geometric techniques~\cite{dou_coverage_2021, lin_point2skeleton_2021, rodrigues_part-based_2018}.
	Our approach to segmentation does not apply any learned components but is fast enough for real-time use by a collaborative robot.
	
	A hierarchical volume-based segmentation technique (HTC) proposed in \cite{attene_hierarchical_2008}, enabled by fast tetrahedralization \cite{si_tetgen_2015} and quick convex hull computation \cite{barber_quickhull_1996}, can perform well if a watertight mesh can be reconstructed from the input point cloud.
	This technique, which we describe in detail in Section \ref{sec:visual_part_segmentation}, is an important element of our part segmentation pipeline.
	Another key component of our approach is the surface-based algorithm proposed in \cite{lin_toward_2018}, which uses a dissimilarity metric to iteratively merge nearby mesh patches, taking special care to preserve part boundaries.

	\section{Inertial Parameters of Homogeneous Parts}
	
	In this section, we describe our inertial parameter identification technique, which assumes that an object has been segmented into its constituent parts.
	By determining the mass of each segment, we are able to identify the full set of inertial parameters, or to provide an approximate solution when \Cref{assum:Homogeneous} is not respected.%
	
	\subsection{Notation}
	Reference frames $\ObjectFrame$ and $\SensorFrame$ are attached to the object and to the force-torque (FT) sensor, respectively.
	The reference frame $\WorldFrame$ is fixed to the base of the robot and is assumed to be an inertial frame, such that the gravity vector expressed in the sensor frame is given by $\Vector{g}_{s} = \Rot{w}{s} [0,0,-9.81]^\Transpose$.
	The orientation of $\ObjectFrame$ relative to $\SensorFrame$ is given by $\Rot{b}{s}$ and the origin of $\ObjectFrame$ relative to $\SensorFrame$, expressed in $\WorldFrame$, is given by $\Pos{b}{s}{w}$.
	The skew-symmetric operator $\Skew{\cdot}$ transforms a vector $\Vector{u} \in\Real^3$ into a $\Real^{3\times3}$ matrix such that $\Skew{\Vector{u}} \Vector{v} = \Vector{u} \times \Vector{v}$.
	
	\subsection{Formulation of the Optimization Problem}
	For a part $\Part{j}$, under \Cref{assum:Homogeneous}, the $k$-th moment of a mass distribution discretized into $n$ point masses is given by
	\begin{equation}
		\label{eqn:Moments}
		\int_{V_j} \Vector{\Position}^k \rho_j(\Vector{\Position}) dV_j \approx \frac{\Mass}{n}\sum_{i}^{n} (\Vector{\Position}_i)^k ~,
	\end{equation}
	where the position of the $i$-th point mass relative to $\ObjectFrame$ is given by $\Vector{\Position}_i$.
	For a homogeneous mass density, the centre of mass corresponds to the centroid
	\begin{equation}
		\Pos{\Part{j}}{b}{b} = \frac{1}{n}\sum_{i}^{n}\Vector{\Position}_i~.
	\end{equation}
	The inertia tensor of the $i$-th point mass relative to $\ObjectFrame$ is
	\begin{align}
		\InertiaMatrix(\Vector{\Position}_i) 
		&= - \Mass_i \Skew{\Vector{\Position}_i} \Skew{\Vector{\Position}_i}\\[1mm]
		&= \Mass_i \begin{bmatrix}
			y^2 + z^2 & -xy & -xz\\
			-yx & x^2+z^2 & -yz\\
			-zx & -zy & x^2+y^2
		\end{bmatrix},
	\end{align}
	where $\Vector{\Position}^{\Transpose}_i = [x, y, z]$ and $\Mass_i$ is the mass of the point.
	
	The part's inertial parameters with respect to the sensor frame $\SensorFrame$ are
	\begin{align}
		\label{eqn:ParamsFromPointMasses}
		&{}^s\Vector{\phi}^{\Part{j}} = \bbm m,\!\! & {}^s\Vector{c}^{\Part{j}},\!\! & {}^s\InertiaMatrix^{\Part{j}} \ebm^\Transpose = \\[1mm]
		&\Mass\! \bbm 
		1\\ 
		\Pos{\Part{j}}{s}{s}\\ 
		\Vech\!\left(\Rot{b}{s} \frac{1}{n}\sum_{i}^{n}\left(-\Skew{\Vector{\Position}_i}\!\Skew{\Vector{\Position}_i}\!\right) \Rot{s}{b} - \Skew{\Pos{\Part{j}}{s}{s}}\! \Skew{\Pos{\Part{j}}{s}{s}}\! \right) 
		\ebm , \notag
	\end{align}
	where $\Rot{b}{s}$ is the rotation that aligns $\ObjectFrame$ to $\SensorFrame$, $\Pos{\Part{j}}{s}{s}$ is the translation that brings the centroid of $\Part{j}$ to $\SensorFrame$, and $\Vech(\cdot)$ is the \emph{vector-half} operator defined in \cite{henderson_vec_1979} that extracts the elements on and above the main diagonal.
	
	By keeping $\Vector{\Position}_i$ fixed and by measuring $\Pos{b}{s}{s}$ and $\Rot{b}{s}$ with the robot's perception system, it becomes clear that only the part's mass $\Mass$ needs to be inferred in \Cref{eqn:ParamsFromPointMasses} as $\Pos{\Part{j}}{s}{s} = \Rot{b}{s}\Pos{\Part{j}}{b}{b}+\Pos{b}{s}{s}$.
	Hence, assuming that the robot's perception system can provide $\Vector{\Position}_i$ and that $\Pos{b}{s}{s}$ and $\Rot{b}{s}$ are either known or measured, the inertial parameters of a homogeneous part depend solely on its mass.
	Similarly, the inertial parameters of a rigid object can be expressed as a function of the masses of its constituent parts.
	
	For ``stop-and-go" motions, where force measurements are taken while the robot is immobile, only the mass and \TextCOM are identifiable \cite{nadeau_fast_2022}.
	Nonetheless, a stop-and-go trajectory greatly reduces noise in the data matrix, because accurate estimates of the end-effector kinematics are not needed \cite{nadeau_fast_2022}.
	Assuming that the manipulated object is built from up to four homogeneous parts\footnote{For stop-and-go trajectories, the rank of the data matrix is four when using non-degenerate poses as described in the appendix of \cite{nadeau_fast_2022}. Dynamic trajectories can increase the rank of $\DataMatrix$ to 10, enabling mass identification of up to 10 unique homogeneous parts.}, finding the mass of each part enables the identification of the complete set of inertial parameters even when stop-and-go trajectories are performed.
	This identification requires measuring the wrench $\Vector{b}_j$ at time step $j$ and relating the wrench to the masses $\Vector{\Mass}$ via
	\begin{equation}
		\underset{\Vector{b}_j}{\underbrace{
				\begin{bmatrix}
					\Vector{\Force}_s\\
					\Vector{\Torque}_s
				\end{bmatrix}
		}}
		=
		\underset{\RedModel_j}{\underbrace{
				\begin{bmatrix}
					\Vector{\Gravity}_s\lbrack1&1&1&1\rbrack\\
					-\lbrack\Vector{\Gravity}_s\rbrack_\times \lbrack\Pos{\Part{1}}{s}{s}&{}\Pos{\Part{2}}{s}{s}&\Pos{\Part{3}}{s}{s}&\Pos{\Part{4}}{s}{s}\rbrack
				\end{bmatrix}
		}}
		\underset{\Vector{\Mass}}{\underbrace{
				\begin{bmatrix}
					m_{\Part{1}}\\m_{\Part{2}}\\m_{\Part{3}}\\m_{\Part{4}}
				\end{bmatrix}
		}},
	\end{equation}
	stacking $K$ matrices such that $\DataMatrix = [\RedModel_1^\Transpose, ..., \RedModel_K^\Transpose]^\Transpose$ and $\Vector{b} = [\Vector{b}_1^\Transpose, ..., \Vector{b}_K^\Transpose]^\Transpose$.
	Minimizing the Euclidean norm leads to the convex optimization problem
	\begin{align}
		\label{eqn:ObjFunc}
		&\min_{\Vector{\Mass} \in\Real^4} \quad\vert\vert \DataMatrix \Vector{\Mass} - \Vector{b} \vert\vert_2 \\
		&\quad\text{\emph{s.t.}} \quad \Mass_{\Part{j}} \ge 0 \enspace \forall j \in \{1, \ldots, 4\}, \notag
	\end{align}
	which can be efficiently solved with standard methods.
	
	\section{Visual Part Segmentation}
	\label{sec:visual_part_segmentation}
	In this section, we combine a part segmentation method that uses local information (e.g., surface normals) with a second method that relies on structural information (e.g., shape convexity).
	The Python implementation of our part segmentation algorithm, which is described by \Cref{algo:WarmStartedHTC}, includes an open-source version of \cite{attene_hierarchical_2008} as well as our variant of \cite{lin_toward_2018} that makes use of colour information.
	
	Defining the shape of an object from a point cloud involves reconstructing the object surface (i.e., shape reconstruction).
	In this work, the ball-pivoting algorithm \cite{bernardini_ball-pivoting_1999} is used owing to its speed and relative effectiveness on point clouds of low density.
	Shape reconstruction can be challenging for objects with very thin parts (e.g., a saw) and a thickening of the object through voxelization is performed beforehand. 
	
	As stated by \Cref{eqn:Moments}, the moments of a mass distribution are computed by integrating over the shape of the distribution. 
	Hence, volumetric part segmentation is sufficient for identification of the true inertial parameters of an object.
	To obtain such a representation of the object from its surface mesh, tetrahedralization is performed via TetGen \cite{si_tetgen_2015} and the resulting tetrahedral mesh is supplied as an input to the part segmentation algorithm.
	
	Our method makes use of the Hierarchical Tetrahedra Clustering (HTC) algorithm \cite{attene_hierarchical_2008}, which iteratively merges clusters of tetrahedra such that the result is as convex as possible while also prioritizing smaller clusters.
	HTC maintains a graph with nodes representing clusters and edges representing adjacency, and with an edge cost based on the concavity and size of the connected nodes. 
	The concavity of a cluster is computed by subtracting its volume from its convex hull and an edge cost is defined by
	\begin{align}
		\label{eqn:htc_cost}
		c_{ij} &= \text{CVXHULL}\left(C_i \cup  C_j\right) - \text{VOL}(C_i \cup  C_j) \notag\\
		\text{Cost}(i,j) &= \begin{cases}
								c_{ij}+1, &\text{if }c_{ij}>0\\
								\frac{\vert C_i\vert^2+\vert C_j\vert^2}{N^2}, &\text{otherwise}
							\end{cases},
	\end{align}
	where $\vert C_i\vert$ is the number of elements in cluster $C_i$ and $N$ is the total number of elements.
	The edge associated with the lowest cost is selected iteratively and the connected nodes are merged into a single cluster, resulting in the hierarchical segmentation.
	
	To make part segmentation faster, we perform an initial clustering such that HTC 
	requires fewer 
	convex hull computations, which is by far the most expensive operation.
	The initial clustering is provided through a bottom-up point cloud segmentation algorithm \cite{lin_toward_2018} that clusters points together based on heuristic features and chooses a representative point for each cluster.
	Two adjacent clusters are merged if the dissimilarity between their representative points is lower than some threshold $\beta$ (n.b.\ the $\lambda$ symbol is used in \cite{lin_toward_2018}). 
	The value of $\beta$ increases with each iteration; the process halts when the desired number of clusters is obtained.
	In our implementation, the dissimilarity metric is defined as:
	\begin{equation}
		\label{eqn:similarity_metric}
		\text{D}(C_i,C_j) = \lambda_p\Norm{\Vector{p}_i-\Vector{p}_j} + \lambda_l\Norm{\Vector{l}_i-\Vector{l}_j} + \lambda_n \left(1-\vert \Vector{n}_i\cdot \Vector{n}_j\vert\right)
	\end{equation}
	where each $\lambda$ is a tunable weight, $\Vector{p}_i$ is the position of the representative point of cluster $C_i$, $\Vector{l}_i \in \Natural^{3}$ is the RGB representation of its colour, and $\Vector{n}_i$ is its local surface normal. 
	
	To enable accurate part segmentation, the initial clustering should not cross the boundaries of the parts that will subsequently be defined by the HTC algorithm.
	Therefore, initial clustering is stopped when the number of clusters (e.g., 50 in our case) is much larger than the desired number of parts.
	The desired number of clusters does not need to be tuned on a per-object basis as long as it is large enough.

	\SetAlgoSkip{SkipBeforeAndAfter}
	\begin{algorithm}[h]
		\DontPrintSemicolon
		\SetKwInOut{Input}{input}\SetKwInOut{Output}{output}
		\Input{A point cloud}
		\Output{A segmented tetrahedral mesh}
		\nlset{1}Initialize similarity threshold $\beta$ as done in \cite{lin_toward_2018}\;
		Initialize each point as a cluster\;
		\While{$number~of~clusters > desired~number$}{
			\ForEach{existing cluster $C_i$}{
				\ForEach{neighboring cluster $C_j$}{
					\If{$\text{D}(C_i,C_j)<\beta$}{
						Merge $C_j$ into $C_i$\;
					}
				}
			}
			$\beta \leftarrow 2\beta$
		}
		\ForEach{point $\Vector{p}$ at the border of two clusters}{
			Merge $\Vector{p}$ into the most similar cluster\;
		}
		\nlset{2}Perform surface reconstruction and tetrahedralization\;
		Associate each tet. to its nearest cluster\;
		Initialize a node in the graph for each cluster\;
		\ForEach{node}{
			\If{two tet. from two nodes share a face}{
				Create an edge between the nodes and compute edge-cost with \Cref{eqn:htc_cost}
			}
		}
		\While{$number~of~edges > 0$}{
			Find edge with lowest cost and merge its nodes\;
			Create a parent node that contains merged nodes\;	
		}
		Label each tet. with its associated cluster number\;
		\caption{HTC (\textbf{2}) with initial clustering (\textbf{1})}
		\label{algo:WarmStartedHTC}
	\end{algorithm}
	
	\section{Experiments}
	In this section, each component of the proposed method is evaluated on 20 objects with standard metrics, and our entire identification pipeline is benchmarked in 80 scenarios.
	The practicality of the proposed approach is also demonstrated through a real `hammer balancing act' experiment using relatively inexpensive robot hardware (see \Cref{fig:demo}). 
	
	To conduct simulation experiments with realistic objects and evaluate the performance of part segmentation and identification, a dataset with ground truth inertial parameters and segments is needed.
	To the best of the authors' knowledge, no such dataset is freely available.
	From CAD files contributed by the community, we built a dataset containing 20 commonly-used workshop tools. 
	For each object, our dataset contains a watertight mesh, a coloured surface point cloud with labelled parts, the object's inertial parameters, and a reference frame specifying where the object is usually grasped.
	This dataset of realistic objects enables the evaluation of shape reconstruction, part segmentation, and inertial parameter identification.

	\subsection{Experiments on Objects from Our Dataset} \label{sec:experiments_on_dataset_objects}
	The quality of a shape reconstructed from point cloud data is evaluated by computing the Hausdorff distance between the ground truth mesh and the reconstructed mesh, both scaled such that the diagonal of their respective bounding box is one metre in length.
	Part segmentation performed with our proposed variation of HTC is evaluated via the undersegmentation error (USE) \cite{levinshtein_turbopixels_2009}, and with the global consistency error (GCE) \cite{martin_database_2001, chen_benchmark_2009}, with results summarized in \cref{tab:dataset_partseg_eval}.
	The USE measures the proportion of points that are crossing segmentation boundaries, or \emph{bleeding out} from one segment to another.
	The GCE measures the discrepancy between two segmentations, taking into account the fact that one segmentation can be more refined than the other.
	Both evaluation metrics represent dimensionless error ratios and are insensitive to over-segmentation, 
	as discussed in Section \ref{sec:Discussion}.
	
	\Cref{tab:AverageComputeTime} summarizes the runtime performance obtained by augmenting HTC with the initial clustering in \Cref{algo:WarmStartedHTC}.
	While the average segmentation error is almost identical in both cases, \Cref{algo:WarmStartedHTC} executes in about a third of the time, owing to the smaller number of convex hull computations performed. 
	\begin{table}[t]
		\centering
		\caption{Average computation time and segmentation error per object from our dataset with standard deviations in parentheses. Initial clustering significantly reduces the runtime with little impact on the part segmentation.}
		\label{tab:AverageComputeTime}
		\begin{tabular}{cccc}
			\toprule
			Algorithm & USE & GCE & Time (s)\\
			\midrule
			HTC                         & 0.1 (0.13) & 0.05 (0.10) & 9.73 (5.56)\\
			HTC with Initial Clustering & 0.1 (0.11) & 0.07 (0.11) & 3.48 (1.00)\\
			\bottomrule
		\end{tabular}
		\vspace{-1mm}
	\end{table}
	\begin{table}[h!]
		\centering
		\caption{Evaluation of part segmentation (USE and GCE) and shape reconstruction (Hausdorff). As expected, objects with a single part do not have GCE and USE errors while objects with many parts have larger segmentation errors.}
		\label{tab:dataset_partseg_eval}
		\begin{tabular}{cccccc}
			\toprule
			Object & USE & GCE & Hausdorff & Mass & \#Parts\\
				   &     &     & (mm)      & (g)  & \\
			\midrule
			Allen Key			&0		&0		&8.2	&128 &1\\
			Box Wrench			&0		&0		&10.9	&206 &1\\
			Measuring Tape		&0		&0		&11.4	&136 &1\\
			Ruler				&0		&0		&14.8	&9 	 &1\\
			Screwdriver			&0.013	&0.001	&12.1	&30  &2\\
			Nut Screwdriver		&0.002	&0.002	&16		&81  &2\\
			Rubber Mallet		&0.011	&0.009	&15.8	&237 &2\\
			Bent Jaw Pliers		&0.176	&0.01	&16.5	&255 &3\\
			Machinist Hammer	&0.018	&0.012	&13.6	&133 &2\\
			Pliers				&0.123	&0.041	&10.7	&633 &3\\
			C Clamp				&0.103	&0.077	&9.6	&598 &5\\
			Adjustable Wrench	&0.117	&0.098	&14.2	&719 &4\\
			Hammer				&0.08	&0.098	&14.7	&690 &3\\
			File				&0.057	&0.102	&17.9	&20  &3\\
			Socket Wrench		&0.141	&0.123	&7.4	&356 &5\\
			Hacksaw				&0.129	&0.128	&11.1	&658 &7\\
			Clamp				&0.259	&0.181	&17.6	&340 &7\\
			Vise Grip			&0.158	&0.287	&17.2	&387 &8\\
			Electronic Caliper	&0.42	&0.316	&9.1	&174 &14\\
			Vise Clamp			&0.296	&0.373	&15.1	&225 &9\\
			\bottomrule
		\end{tabular}
		\vspace{-5mm}
	\end{table}
	\begin{figure}
		\vspace{-3mm}
		\centering
		\begin{overpic}[width=1\linewidth]{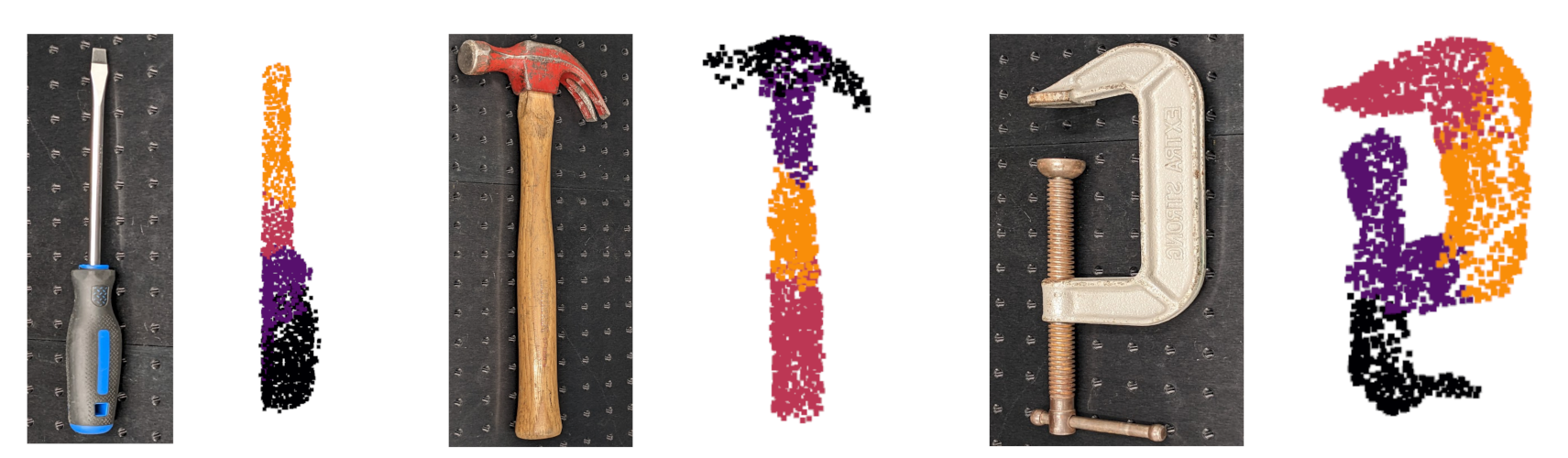}
			\put(12,-1){(a)}
			\put(43,-1){(b)}
			\put(82.5,-1){(c)}
		\end{overpic}
		\vspace{-3mm}
		\caption{Pictures of objects scanned by the RGB-D camera on the manipulator, next to their segmented meshes; points are located at the tetrahedra's centroids (screwdriver is rotated).}
		\label{fig:realsegvspictures}
	\end{figure}
	The reconstruction and part segmentation can also be qualitatively evaluated on point clouds obtained from RGB-D images taken by a Realsense D435 camera at the wrist of the manipulator.
	To complete a shape that is partly occluded by the support table, the point cloud is projected onto the table plane, producing a flat bottom that might not correspond to the true shape of the object. For instance, the left side of the rotated screwdriver point cloud in \Cref{fig:realsegvspictures} is the result of such a projection.
	
	\begin{table}[t]
		\centering
		\caption{Standard deviations of the zero-mean Gaussian noise added to accelerations and force signals in simulation.}
		\label{tab:noise_std_dev}
		\begin{tabular}{ccccc}
			\toprule
			Scenario  		& Ang. Acc. & Lin. Acc. & Force & Torque\\
			\midrule
			Low Noise 		& 0.25  & 0.025 & 0.05 & 0.0025\\
			Moderate Noise 	& 0.5   & 0.05  & 0.1  & 0.005\\
			High Noise 		& 1		& 0.1   & 0.33 & 0.0067\\
			\bottomrule 
		\end{tabular}
	\end{table}
	
	\begin{figure}
		\vspace{-1mm}
		\centering
		\includegraphics[width=1\linewidth]{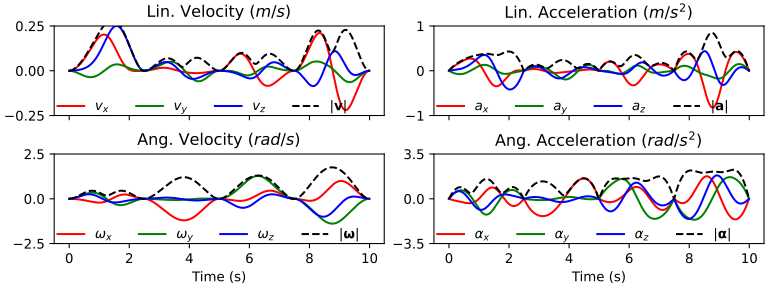}
		\caption{Kinematics of the identification trajectory performed with the Rubber Mallet. The norm of the velocity (black) goes slightly above the standard maximum speed for a cobot.}
		\label{fig:kinematicsrubbermallet}
		\vspace{-5mm}
	\end{figure}
	
	The proposed identification algorithm is tested in simulation under four noise scenarios where zero-mean Gaussian noise is added to the signals. The  standard deviations of the noise values are based on the specifications of the Robotiq FT-300 sensor as described in \Cref{tab:noise_std_dev}.
	The identification trajectory used to generate the regressor matrix $\DataMatrix$ in \Cref{eqn:ObjFunc} has the robot lift the object and successively orient it such that gravity is aligned with each axis of $\ObjectFrame$, stopping for a  moment at each pose and moving relatively quickly between poses, as shown in \Cref{fig:kinematicsrubbermallet}.
	The average condition number~\cite{gautier_exciting_1992} of the scaled regressor matrix for all simulated trajectories is 74, 92, and 99 for the low, moderate, and high noise scenarios, respectively.
	These relatively low condition numbers confirm that our identification trajectory is non-degenerate.

	The performance of our proposed algorithm (HPS) is compared against the classical algorithm proposed in \cite{atkeson_estimation_1986} (OLS) and to the more modern algorithm proposed in \cite{lee_geometric_2019} (GEO). 
	The latter was provided with a prior solution consisting of the true mass of the object, and the \TextCOM and inertia tensor resulting from a homogeneous mass distribution for the object ($\alpha=10^{-5}$ was used).
	HPS only uses measurement data when
	both the linear and angular accelerations are below 1 unit$/s^2$,
	corresponding to the \textit{stalled} timesteps of the stop-and-go motion, whereas OLS and GEO use all data available.
	
	The accuracy of inertial parameter identification is measured via the Riemannian geodesic distance ${e_{\text{Rie}} = \sqrt{\left(\frac{1}{2}\sum\Vector{\lambda}\right)}}$~\cite{lang_fundamentals_1999}, where $\Vector{\lambda}$ is the vector of eigenvalues for $P({}^s\phi^1)^{-1}P({}^s\phi^2)$ and $P(\phi)$ is the \textit{pseudoinertia} matrix that is required to be symmetric positive-definite (SPD) \cite{wensing_linear_2017}.
	The metric $e_{\text{Rie}}$ is the distance between the estimated and ground truth inertial parameters in the space of SPD matrices.
	To assist interpretation of the identification error, we also compute the size-based error metrics proposed in \cite{nadeau_fast_2022}, which use the bounding box and mass of the object to  produce average \textit{percentage} errors $\bar{\Vector{e}}_m$, $\bar{\Vector{e}}_C$, and $\bar{\Vector{e}}_J$ associated, respectively, with the mass, centre of mass, and inertia tensor estimates.
	\Cref{tab:PerfoComparison} reports the mean of the entries of the error vector $\bar{\Vector{e}}$ for each quantity of interest. 
	
	\begin{table}[t]
		\centering
		\caption{Comparison between HPS (ours), OLS, and GEO with various levels of noise, indicating the percentage of solutions that were physically consistent (Cons).}
		\label{tab:PerfoComparison}
		\begin{tabular}{ccccccc}
			\toprule
			Noise 	& Algo. & Cons. (\%) & $\bar{\Vector{e}}_m$(\%) & $\bar{\Vector{e}}_C$(\%) & $\bar{\Vector{e}}_J$(\%) & $e_{\text{Rie}}$\\
			\midrule
			No	 & OLS & 100 & \textbf{$<$0.1} & \textbf{$<$0.1} & \textbf{0.09}    & \textbf{0.03}\\
				 & GEO & 100 & $<$0.1 & 1.35   & 53.22	 & 1.14\\
				 & HPS & 100 & 0.27   & 0.1    & 10.28   & 0.72\\
			Low  & OLS & 14  & 0.19   & 2.13   & $>$500  & N/A\\
				 & GEO & 100 & \textbf{0.18}	  & 1.34   & 52.79   & 1.14\\
				 & HPS & 100 & 0.40   & \textbf{0.32}   & \textbf{11.12}   & \textbf{0.74}\\
			Mod. & OLS & 4.5 & 0.36   & 5.33   & $>$500  & N/A\\
				 & GEO & 100 & \textbf{0.35}   & 1.37   & 51.73   & 1.14\\
				 & HPS & 100 & 0.74   & \textbf{0.48}   & \textbf{11.81}   & \textbf{0.77}\\
			High & OLS & 2   & 0.69   & 10.11  & $>$500  & N/A\\
			 	 & GEO & 100 & \textbf{0.64}   & 1.58   & 48.49   & 1.13\\
				 & HPS & 100 & 2.79   & \textbf{1.07}   & \textbf{15.00}   & \textbf{0.87}\\
			\bottomrule
		\end{tabular}
		\vspace{0mm}
	\end{table}

	\subsection{Demonstration in Real Settings}
\begin{figure}[t]
		\centering
		\begin{overpic}[width=1\linewidth]{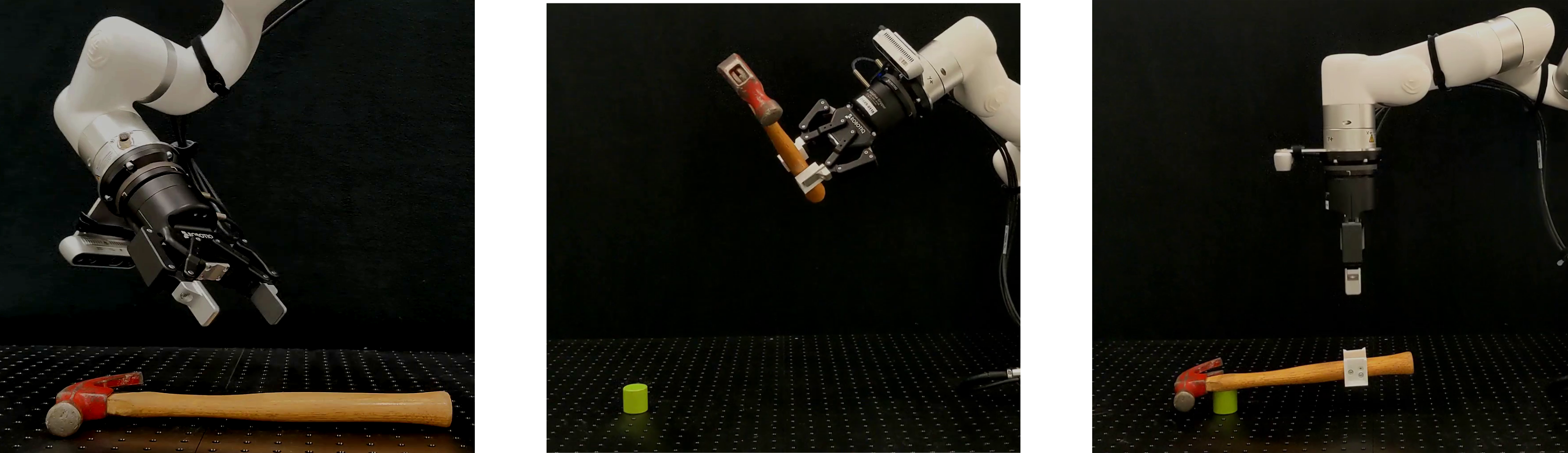}
			\put(12,-4){(a)}
			\put(47,-4){(b)}
			\put(82.5,-4){(c)}
		\end{overpic}
		\vspace{0.1mm}
		\caption{The hammer is (a) scanned, (b) picked up for inertial identification, and (c) balanced onto the small green target.}
		\label{fig:demo}
		\vspace{-4mm}
	\end{figure}

	To test our proposed method in a realistic setting, we used a uFactory xArm 7 manipulator equipped with a RealSense D435 camera and a Robotiq FT-300 force-torque sensor, as shown in \Cref{fig:gladyswieldingobject}.
	First, the hammer in \Cref{fig:realsegvspictures} was scanned with the camera, producing 127 RGB-D images of the scene in about 30 seconds.
	The object was then picked up at a predetermined grasping pose where the Robotiq 2F-85 gripper could fit into plastic holders attached to the object (enabling a stable grasp of the handle).
	A short trajectory that took about 10 seconds to execute was performed while the dynamics of the object were recorded at approximately 100 Hz.
	Point cloud stitching, mesh reconstruction, and part segmentation can be  performed concurrently while the robot executes the trajectory, since these operations take 2.87, 0.24, and 2.82 seconds, respectively.
	Finally, our proposed method identified the inertial parameters in about 0.5 seconds with MOSEK \cite{andersen2000mosek}. 
	Using only its estimate of the \TextCOMMA, the robot autonomously balanced the hammer on a cylindrical target with a radius of 17.5 mm.
	The entire process is shown in our accompanying video, with summary snapshots provided in \Cref{fig:demo}.
	In contrast, both OLS and GEO returned inaccurate parameter estimates, causing hammer balancing to fail.
	
	\section{Discussion}
	\label{sec:Discussion}
	This work exploits object shape information to produce fast and accurate inertial parameter estimates.
	The object reconstructions used by our method to determine object shape are often already computed as components of planning and manipulation algorithms, limiting the computational burden introduced by our approach.
	The decomposition of objects into parts also enables fast inertial parameter identification of bodies that have joints (e.g., hedge scissors), since the parameters can be trivially updated following a change in object configuration.
	
	Wrong or inaccurate part segmentation may lead to erroneous parameter estimates.
	However, if \Cref{assum:Homogeneous} holds, \emph{over}-segmenting an object does not affect the result of the identification as for a given part, \Cref{eqn:Moments} can be decomposed into
	\begin{equation}
		\label{eqn:Oversegmentation}
		 \int_{V_1} \Vector{\Position}^k \rho_1(\Vector{\Position}) dV_1 + \int_{V_2} \Vector{\Position}^k \rho_2(\Vector{\Position}) dV_2 = \int_{V} \Vector{\Position}^k \rho(\Vector{\Position}) dV,
	\end{equation}
	which is true since $V=V_1 \cup V_2$ and $\rho_1(\Vector{\Position}) = \rho_2(\Vector{\Position})$.
	Similarly, if two conceptually distinct parts with the \emph{same} mass density are erroneously combined into one, the result of the identification will remain unaffected.
	However, if parts with \emph{different} mass densities are considered to be a single part by the algorithm, the identification will fail since \Cref{eqn:Oversegmentation} does not hold when $\rho_1(\Vector{\Position}) \neq \rho_2(\Vector{\Position})$.
	
	The comparison in \Cref{tab:PerfoComparison} suggests that OLS outperforms other algorithms in the noiseless scenario, which is expected since it is not biased by any prior information.
	However, OLS nearly always converges to physically inconsistent solutions and becomes inaccurate in the presence of even a small amount of sensor noise.
	The GEO algorithm performs similarly across noise levels, possibly due to the very good prior solution (i.e., correct mass, homogeneous density) provided, corresponding to the exact solution for objects that have a single part. 
	On average, HPS outperforms OLS and GEO for the identification of \TextCOM and $\InertiaMatrix$ when the signals are noisy.
	The slightly higher $\bar{\Vector{e}}_m$ for HPS may be caused by the approximation made when using stalling motions~\cite{nadeau_fast_2022}.
	
	Experiments with objects from our dataset do not reveal any obvious trends relating the quality of the shape reconstruction (measured via the Hausdorff distance), the quality of the part segmentation (measured via USE and GCE), and the quality of the inertial parameter identification (measured via $e_{\text{Rie}}$).
	This can be explained by the fact that an object's mass and shape largely determine the signal-to-noise ratios that any identification algorithm has to deal with.
	
	As demonstrated by experiments with objects that are mostly symmetrical (e.g., the screwdriver), if the shape of the object is such that the parts' centroids are coplanar, the optimizer will `lazily' zero out the mass of some parts since they are not required to minimize \Cref{eqn:ObjFunc}.
	An improved version of HPS could use the hierarchy from HTC to intelligently define segments whose centroids are not coplanar.
	
	\section{Conclusion}
	\label{sec:Conclusion}
	
	In this paper, we leveraged the observation that man-made objects are often built from a few parts with homogeneous densities. 
	We proposed a method to quickly perform part segmentation and inertial parameter identification.
	We ran 80 simulations in which our approach outperformed two benchmark methods, and we demonstrated real-world applicability by autonomously balancing a hammer on a small target. 
	On average, our proposed algorithm performs well in noisy conditions and can estimate the full set of inertial parameters from`stop-and-go' trajectories that can be safely executed by collaborative robots.
	Promising lines of future work include formulating the optimization problem as a mixed-integer program where mass densities are chosen from a list of known materials, and improving the segmentation algorithm such that parts' centroids are never coplanar.
	
	\bibliographystyle{ieeetr}
	\bibliography{Extracted}
\end{document}